%% file: main.tex
\pdfoutput=1

\documentclass[11pt]{article}

\usepackage[final]{acl}

\usepackage{times}
\usepackage{latexsym}

\usepackage[T1]{fontenc}

\usepackage[utf8]{inputenc}

\usepackage{microtype}

\usepackage{inconsolata}

\usepackage{graphicx}

\usepackage{algorithm}
\usepackage{algorithmic}

\usepackage{multirow}
\usepackage{csquotes}
\usepackage{subcaption}
\usepackage[T1]{fontenc}
\usepackage{xcolor}

\usepackage{newfloat}
\usepackage{listings}

\newcommand{\ignore}[1]{}

\title{\includegraphics[scale=0.04]{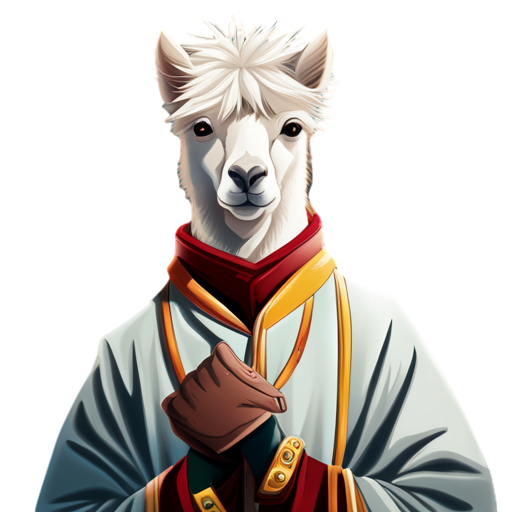}
\texttt{SCITUNE:} Aligning Large Language Models with Human-Curated Scientific Multimodal Instructions}

\author{Sameera Horawalavithana, Sai Munikoti, Ian Stewart, Henry Kvinge, Karl Pazdernik \\
        Pacific Northwest National Laboratory, Richland, WA}
        
\begin{document}

\maketitle

\begin{abstract}
Instruction finetuning is a popular paradigm to align large language models (LLM) with human intent. 
Despite its popularity, this idea is less explored in improving LLMs to align existing foundation models with scientific disciplines, concepts and goals. 
In this work, we present \textit{SciTune} as a tuning framework to improve the ability of LLMs to follow multimodal instructions generated from scientific publications. 
To test our methodology, we train a large multimodal model LLaMA-SciTune that connects a vision encoder and LLM for science-focused visual and language understanding.  
LLaMA-SciTune significantly outperforms the state-of-the-art models in the generated figure types and captions in SciCap and VisText benchmarks. 
In comparison to the models that are finetuned with synthetic data only, LLaMA-SciTune surpasses human performance on average and in many sub-categories on the ScienceQA benchmark.
Our results demonstrate that human-generated scientific multimodal instructions remain highly valuable in tuning LLMs to perform well on science tasks, despite their lower volume and relative scarcity compared to synthetic data.
We publicly release the SciTune codebase\footnote{\url{https://github.com/pnnl/scitune}}.
\end{abstract}

\input{PNNL-ST-SciTune/sections/intro}
\input{PNNL-ST-SciTune/sections/methodology}

\input{PNNL-ST-SciTune/sections/experiments}
\input{PNNL-ST-SciTune/sections/related}
\input{PNNL-ST-SciTune/sections/conclusion}

\input{PNNL-ST-SciTune/sections/ack}



\bibliography{bib-v1}

\appendix
\input{PNNL-ST-SciTune/sections/appendix}

\end{document}

%% file: PNNL-ST-SciTune/sections/intro.tex
\section{Introduction}




Instruction finetuning has gained significant traction in the NLP community as a means of enhancing the capabilities of large language models (LLMs), allowing them to accurately balance desired outcomes, context, and human preferences, leading to more relevant and coherent responses.
More recently, AI assistants have been trained to comprehend and execute multimodal vision-and-language instructions, aligned with human intent, to accomplish diverse real-world tasks in complex multimodal environments.
In one of the latest developments, MiniGPT-4~\citep{zhu2023minigpt}, LLaVA~\citep{liu2023visual} and LLaMA-Adapter~\citep{gao2023llamaadapter} have focused on expanding language-only instruction models to incorporate multimodal capabilities, thereby granting LLMs the ability to perform visual grounded reasoning tasks.

Recent research suggests that high-quality multimodal pretraining data and instructions, such as high-resolution images and diverse multimodal data are important for effective model performance~\citep{mckinzie2024mm1}.
However, one of the primary bottlenecks is the scarcity of high-quality data for multimodal pretraining and instruction tuning. 
To mitigate the challenges associated with data scarcity, many recent multimodal models rely on synthetically generated instructions (training data distilled from responses from other models) for fine-tuning instead of human annotations~\citep{liu2023visual,cascante2023going,bai2022constitutional}.

However, using synthetic data to align AI models can create confusion and uncertainty, since synthetic data, being artificially generated, often fails to capture the complexities of human values~\citep{liu2024best,zhou2024real}. 
This can cause AI models to learn from biased~\citep{feng2023pretraining,liu2021mitigating}, ungrounded~\citep{liu2022mind,patel2022mapping} or inaccurate data~\citep{ji2023survey,weidinger2021ethical}. 
Additionally, training models on recursively generated data can lead to a loss of true data distribution, resulting in less varied and misaligned outputs~\citep{shumailov2024ai}. 
Furthermore, models tuned with synthetic data often fail to meet the standards required by certain scientific subdomains, such as medicine~\citep{li2023llavamed,xia2024cares}.
For example, several open-source medical vision-language models such as LLaVA-Med~\citep{li2023llavamed}, MedFlamingo~\citep{moor2023med}, MedVInT~\citep{zhang2023pmc}, and RadFM~\citep{wu2023towards} failed to satisfy the trustfulness, fairness, safety, privacy, and robustness conditions in the recently introduced CARES~\citep{xia2024cares} benchmark.
As a result, AI systems relying on synthetic data might behave unpredictably and could potentially cause unintended or harmful outcomes~\citep{anderljung2023frontier,zou2023universal}.

We believe this is mainly due to the absence of alignment methods designed to synchronize existing foundation models with scientific disciplines, concepts, and goals and to ensure that the generated content meets the standards and expectations of the scientific community.
Our hypothesis is that scientifically aligned multimodal models can learn from unique patterns and structures present in scientific language generated by human scientists, thus would be able to follow precise instructions about complex procedures, protocols, and guidelines in the scientific environments.
This leads to the question:~\textit{To what extent, can LLMs align solely with human-curated scientific multimodal instructions?}

To this end, we build on top of the LLaVA~\citep{liu2023visual} model architecture to perform scientific multimodal instruction tuning (\textit{SciTune}) on top of a decoder-based pretrained LLM and vision encoder.
The \textit{SciTune} training method includes two stages for \textit{scientific multimodal instruction tuning}, i) \textit{scientific concept alignment} to learn across various scientific visual signals (e.g., plots, charts, equation, diagram, etc.), and textual signals (e.g., captions, optical character recognition (OCR) and
paragraph mentions), ii) \textit{scientific instruction tuning} to fine-tune on a multimodal scientific reasoning task.
To validate our approach, we train our models on top of LLaMA~\citep{touvron2023llama} and the CLIP~\citep{radford2021learning} vision encoder model.
We show that our model surpasses human performance on the ScienceQA multimodal reasoning benchmark and performs significantly better than state-of-the-art vision-language models in a variety of scientific image understanding tasks.
Our results demonstrate that human-curated scientific multimodal instructions remain highly valuable in tuning LLMs to perform well on science tasks, despite their lower volume and relative scarcity compared to synthetic data.

\begin{figure*}[!t]
    \centering
    \includegraphics[scale=0.15]{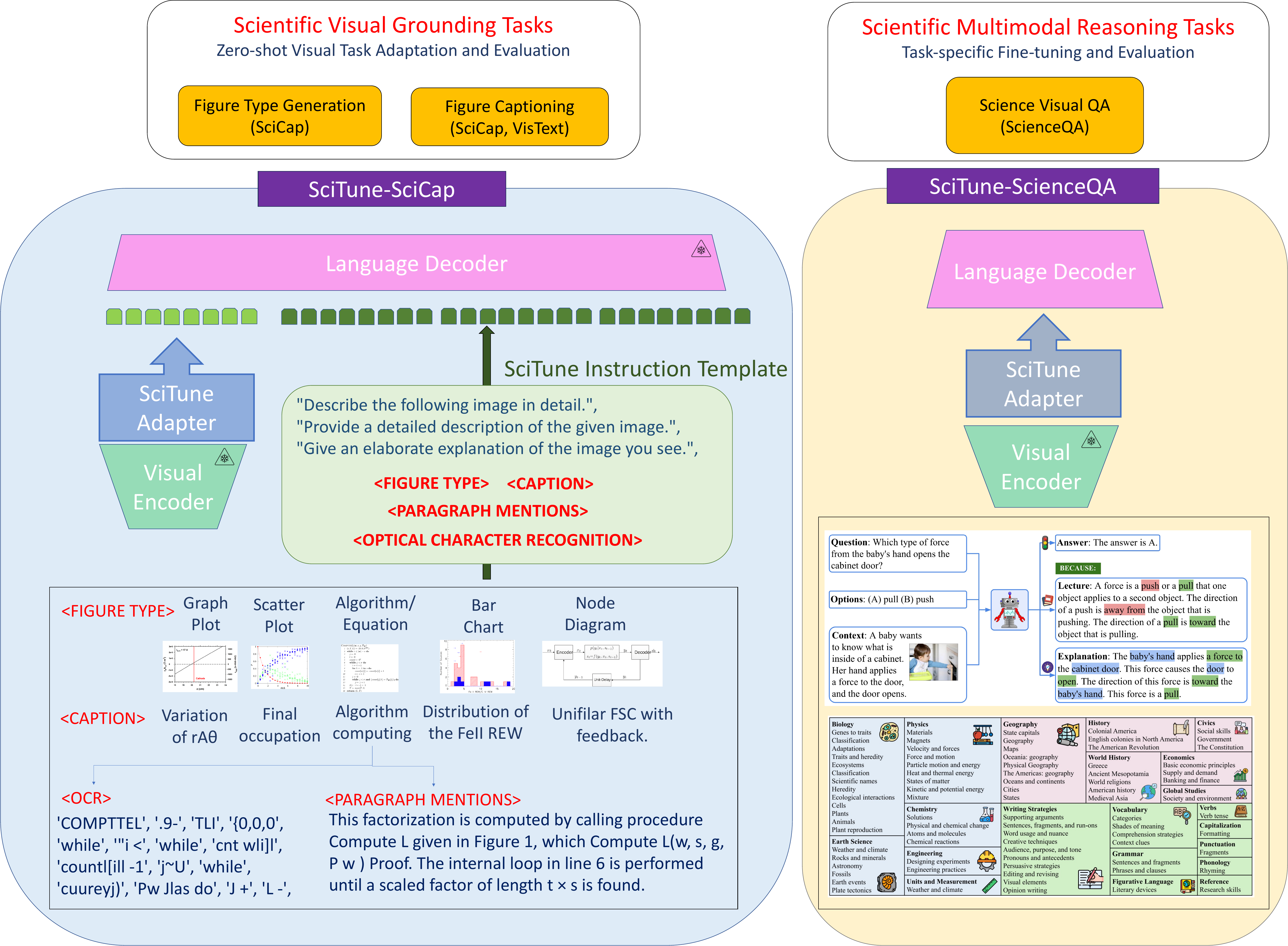}
    \caption{SciTune enables models to learn \textit{scientific concept alignment} across various visual signals (e.g., plots, charts, equation, diagram), and textual signals (e.g., captions, OCR and paragraph mentions); shown in the left graphic in the figure. After alignment, the model can be used to perform various scientific visual grounding tasks (e.g., figure type generation, captioning) with zero demonstrations at inference time (shown in the upper-left graphic). The pretrained model can be further finetuned on a multimodal scientific reasoning task (e.g., ScienceQA; shown in the upper-right graphic).
    }
    \label{fig:scitune_framework}
\end{figure*}

%% file: PNNL-ST-SciTune/sections/methodology.tex
\section{Methodology}

In this section, we describe the SciTune framework in two stages of Scientific Multimodal Concept Alignment and Multimodal Task-specific Instruction Tuning 
and the design choices and multimodal architecture used for the experiments.


\subsection{Scientific Multimodal Instructions}
\label{sec:sci_concept_alignment}

We use an \textit{early-fusion} strategy~\citep{liu2023visual} to jointly reason over the text, images, and other modalities with a shared multifaceted representation as presented as SciTune instructions.
The SciTune instruction template $x=(s_D, s_I, s_{T})$ includes a system message $s_D$ to help the model to understand the role and context, instruction $s_I$ randomly sampled from the visual-grounded questions, and $s_{T}$ to encode the multimodal data.

\paragraph{\textbf{Human-curated Scientific Instructions}}
This work solely focuses on multimodal instructions curated by humans instead of machine generated content used in other visual instruction tuned models~\citep{liu2023visual,gao2023llamaadapter}.
Our goal is to align the pretrained foundation models with natural scientific concepts and the intentions of humans (scientists).
To this end, we chose scientific publications (PDFs) as the medium of scientific instructions that demonstrate various stages of scientific discovery.

We use the SciCap~\citep{hsu2021scicap} dataset with more than 400,000 scientific figure images extracted from various arXiv papers, including their respective captions and relevant paragraphs.
This dataset is composed of arXiv papers from January 2010 to October 2020. 
It consists eight distinct categories: Computer Science, Economics, Electrical Engineering and Systems Science, Mathematics, Physics, Quantitative Biology, Quantitative Finance, and Statistics.
We use the 333,472 examples provided in the SciCap training split for pretraining and use the validation split to evaluate the performance.

We introduce scientific \textbf{c}aptions ($s_c$), figure \textbf{t}ypes ($s_t$), \textbf{o}ptical character recognition (OCR)($s_o$) and
paragraph \textbf{m}entions($s_m$) in the instruction template ($s_{T}=\{s_c,s_t,s_o,s_m\}$) to convert the SciCap dataset into a multimodal instruction-tuning dataset.
Figure-captioning ($s_c$) data typically includes brief text that is highly specific to the associated figure. 
In contrast, interleaved data such as paragraph \textbf{m}entions ($s_m$) generally features longer and more varied text, which is broadly relevant to the figures it associates.
Please see the Appendix (Table~\ref{tab:scitune_example}) for a SciTune instruction sample.

\subsection{Multimodal Architecture}
\label{sec:multimodal_architecture}

\paragraph{\textbf{Architecture}}
We build on top of the most recent multimodal architectures (e.g., LLaVA~\citep{liu2023visual}, LLaMA-Adapter~\citep{zhang2023llama}) that guide LLMs to follow multimodal instructions. 
We noticed that adapter-based multimodal training serves as the most efficient technique for injecting multimodal knowledge to a pretrained LLM decoder model.
Our goal was to improve the existing LLMs to perform better on science-focused multimodal reasoning and visual grounded tasks.
To this end, we chose LLaMA-1~\citep{touvron2023llama} as the LLM decoder, and CLIP visual encoder~\citep{radford2021learning} to experiment with multimodal adapter training as shown in Figure~\ref{fig:scitune_framework}.

The SciTune adapter transforms the output of the visual encoder model as inputs to the language decoder with a linear projection layer.
While we keep the language decoder and the visual encoder models frozen, the multimodal adapter is updated during the pretraining stage.
This modular architecture can be filled by any language decoder and a visual encoder model.
We conduct the experiments with LLaMA 7B and 13B model variants for better comparison with other baseline models.
It is worthwhile to note that we chose LLaMA due to its superior performance in the public benchmarks and its open-source accessibility.

We do not use any instruction-tuned LLaMA variants (e.g., Vicuna, Guanaco) in our experiments due to two main reasons.
First, we mainly focus on improving the base LLM decoder models with multimodal instructions generated by humans in order to eliminate all confounding factors such as machine generated instruction tuning.
Since a majority of instruction-tuned models developed on top of LLaMA are knowledge-distilled from closed-source, proprietary models like GPT-4, we want to avoid any unexpected performance advantages.
Second, we want to make a fair comparison with other baseline models proposed in this area developed on top of the base LLaMA model, and test whether the multimodal instruction tuning proposed in this work could lead into better scientific concept understanding compared to those models.


\paragraph{\textbf{Training}}
We model the language distribution $p(x)$ from a set of SciTune instructions ($x_1,x_2,..,x_m$)
as the product of conditional multimodal token probabilities
as shown in Equation~\ref{eq:estimation}.

\begin{equation}
\label{eq:estimation}
    p(x)=\prod_{j=1}^{n}  p(s_{T>j}|s_V, s_{I},s_{T<j})
\end{equation}

We model $s_V$ with the multimodal tokens after projection from the respective plot visuals $V$.
We sample the instruction $s_I$ from the list of questions presented in the Appendix (Table~\ref{tab:llava_instructions}). 
Note that we skip the token descriptors in $s_{T}$ for brevity, unless the model is trained autoregressively to generate exact tokens across all textual modalities in $s_{T}=\{s_c,s_t,s_o,s_m\}$.
More importantly, the model is able to jointly generate all modality tokens in a single-turn conversation.
For example, given a scientific plot and an instruction, the model first generates the figure type (e.g., Graph Plot, Scatterplot, Node Diagram, Equation, Bar Chart), then the visual content through captioning and OCR, and finally the cited paragraph.

As presented in Figure~\ref{fig:scitune_framework}, \textit{LlaMA-SciTune-Scicap} is tuned to align the LLM towards scientific concepts. 
~\textit{LlaMA-SciTune-Scicap} can be further finetuned on a multimodal scientific reasoning task.
In our experiments, we name the task finetuned model variant as \textit{LlaMA-SciTune-ScienceQA}.

%% file: PNNL-ST-SciTune/sections/experiments.tex
\section{Experiments}
In this section, we report the performance of \textit{LLaMA-SciTune} models across a variety of science-focused downstream tasks.
Our goal is to assess the performance of the models in visual grounded language understanding and multimodal reasoning tasks.
For example, we want to show how much difference the training stages contribute to the model performance, or whether adding various scientific modalities in the instruction template 
improves the overall performance.
Note that our objective is not to introduce a model that tops the leaderboard across various downstream tasks. 
Instead, we aim to demonstrate the extent to how human-curated scientific multimodal instructions can be useful in aligning vision-language models.

To this end, we trained three \textit{LLaMA-SciTune-SciCap} models and finetuned them with scientific multimodal reasoning dataset (e.g., ScienceQA) for the corresponding \textit{LLaMA-SciTune-ScienceQA} models.
Three \textit{LLaMA-SciTune-SciCap} models differ on the text input types (e.g., Caption, Figure Type, OCR, and Figure Mentions) and the scale of the LLM (e.g., 7B and 13B) used in the model training.
For example, \textit{LLaMA-SciTune-SciCap-13B (CTOM)} model uses the base LLaMA-1 13B checkpoint and figure \textbf{c}aption, figure \textbf{t}ype, \textbf{O}CR, and figure \textbf{m}entions in the training.
Similarly, \textit{LLaMA-SciTune-ScienceQA-13B (CTOM)} model is finetuned on the \textit{LLaMA-SciTune-SciCap-13B (CTOM)} with the ScienceQA~\citep{lu2022learn} training split.
While \textit{LLaMA-SciTune-SciCap} model tunes the LLM to better understand scientific concepts, \textit{LLaMA-SciTune-ScienceQA} is further finetuned for scientific reasoning.

First, we report the performance of \textit{LLaMA-SciTune-SciCap} in two science-focused visual grounded tasks to assess the scientific concept alignment training stage (Section~\ref{sec:viz_grounded_tasks}).
Finally, we use the ScienceQA benchmark to test the multimodal reasoning abilities of \textit{LLaMA-SciTune-ScienceQA} across three scientific subject areas (Section~\ref{sec:scienceqa_performance}).


\subsection{Vision Grounded Tasks Performance}
\label{sec:viz_grounded_tasks}
In this section, we report the performance of the \textit{LLaMA-SciTune-SciCap} model for two zero-shot downstream tasks.
Note that, we reference the \textit{LLaMA-SciTune-SciCap-13B (CTOM)} model in this performance analysis.
In the first task, we evaluate how well the \textit{LLaMA-SciTune-SciCap} model is able to align the associated figure types with the actual image.
In the second task, we evaluate the performance of the \textit{LLaMA-SciTune-SciCap} model in generating the figure captions.

\subsubsection{Scientific Figure Type Generation}
In the scientific concept alignment stage, one of the learning tasks is to align the scientific visuals with the correct figure type.
For example, the model should be able to distinguish a graph plot from a scatter plot.
We compare the performance of our model of generating the figure types with a standalone vision encoder.
For example, we use the CLIP model~\citep{radford2021learning} to perform figure type classification in the zero-shot manner given five candidate types (e.g., Graph Plot, Scatterplot, Node Diagram, Equation, Bar Chart).
We use the validation data released by the SciCap challenge 
to perform our experiments.
This validation dataset includes plots and the associated figure types.
We locate the figure types in the generated SciTune outputs, and compare it with the ground truth.
As shown in Table~\ref{tab:type_clf}, \textit{LLaMA-SciTune-SciCap} shows 57\% performance improvement over the standalone CLIP model used in the figure type classification.

It is important to note that the \textit{LLaMA-SciTune-SciCap} used the same CLIP model as the visual encoder, but the additional multimodal adapter was optimized towards aligning figure types with the plots during the pretraining stage.
This multimodal adapter is able to project the outputs of vision encoder into the LLM to improve its understanding on the scientific plots.
One could argue that a more ideal comparison would be between the \textit{LLaMA-SciTune-SciCap} model and a version of the CLIP model that is specifically tuned with the same dataset, rather than comparing it with the vanilla (untuned) CLIP model.
In this experiment, our objective was to evaluate how much the SciTune adapter (Figure~\ref{fig:scitune_framework}) contributes to improving the LLM's performance in understanding scientific plots, as opposed to improvements gained merely through the use of the tuned visual encoder.



\begin{table}[htbp]
    \centering
    \caption{Accuracy of Generating the Figure Types. We also report the zero-shot figure type classification performance of the CLIP model.}
    \label{tab:type_clf}
    \begin{tabular}{|c|r|r|}
    \hline
    Figure Type & CLIP & SciTune-SciCap\\ \hline \hline
         Graph Plot & 52.58 & \textbf{93.63} \\ \hline
         Scatterplot & 52.20 & \textbf{70.14}\\ \hline
         Node Diagram & 77.67 & \textbf{95.40}\\ \hline
         Equation & 60.47 & \textbf{89.54}\\ \hline
         Bar Chart & 32.67 & \textbf{80.33}\\ \hline \hline
         All & 55.11 & \textbf{85.81}\\ \hline
    \end{tabular}
\end{table}

\subsubsection{Scientific Figure Captioning}
In this section, we test the model performance of generating scientific figure captions given only the scientific plot.
Previous works show that scientific figure captioning is an extremely challenging task due to complex image understanding required in vision-to-language modeling~\citep{huang2023summaries}.
We take the first sentence in the generated SciTune output as the generated caption.
We compare~\textit{LLaMA-SciTune-SciCap} model performance with the SOTA image captioning model, BLIP~\citep{li2022blip}, trained with more than 14M image-text pairs.
We use two text evaluation metrics, BLEU and ROUGE, to measure the quality of generated captions with respect to the ground truth captions.
We evaluate the models in two scientific image captioning benchmarks, SciCap and VisText~\citep{tang2023vistext}.
We used the validation split with 47639 and 1202 images in two benchmarks, respectively. 

As shown in Table~\ref{tab:evaluation_figure_captioning}, the \textit{LLaMA-SciTune-SciCap}  model outperforms the BLIP model in both automated text evaluation metrics.
This suggests that \textit{LLaMA-SciTune-SciCap} may have a better understanding of the scientific plot in comparison to the BLIP model finetuned towards image captions.
Table~\ref{tab:gen_caps} (see Appendix) shows a few generated captions in comparison to the baseline and ground truth image captions.

\begin{table}[htbp]
\caption{In-distribution (SciCap) and out-of-distribution (VisText) Evaluation of Generated Figure Captions}
\label{tab:evaluation_figure_captioning}
\centering
\scalebox{0.85}{
\begin{tabular}{|c|c|c|c|}
\hline
Benchmark & Model & BLEU          & ROUGE         \\ \hline \hline
\multirow{2}{*}{SciCap} & BLIP & 0.02$\pm$0.02 & 0.11$\pm$0.07 \\ \cline{2-4}
& SciTune-SciCap & \textbf{0.05$\pm$0.03} & \textbf{0.13$\pm$0.08} \\ \hline
\multirow{2}{*}{VisText} & BLIP & 0.06$\pm$0.05 & 0.23$\pm$0.11 \\ \cline{2-4}
& SciTune-SciCap & \textbf{0.10$\pm$0.07} & \textbf{0.23$\pm$0.12} \\ \hline
\end{tabular}
}
\end{table}

\subsection{Scientific Multimodal Reasoning Task Performance}
\label{sec:scienceqa_performance}
In this section, we evaluate the model performance on science-focused multimodal reasoning question and answering (QA).
We report the \textit{LLaMA-SciTune-ScienceQA} model performance in the ScienceQA benchmark~\citep{lu2022learn} that includes 21k multimodal multiple choice questions with rich domain diversity across 3 subjects, 26 topics, 127 categories, and 379 skills.
We use the ScienceQA training split (12726 examples) to tune the \textit{LLaMA-SciTune-SciCap} model further as shown in Figure~\ref{fig:scitune_framework}.
Table~\ref{fig:sqa_results} reports the performance of the models on the ScienceQA test split (4241 test questions).
While lectures are shared between training and test splits, there are new questions associated with multimodal contexts, and explanations in the test split. 
We have three main observations from this table.
\input{PNNL-ST-SciTune/sections/scienceqa_baseline_results}

First, \textit{LLaMA-SciTune-ScienceQA-13B (CTOM)} model outperforms the human performance on average and in four other sub-groupings.
For example, this model records 90.03\% accuracy in correctly answering the multimodal reasoning questions in the ScienceQA benchmark, where humans record only 88.40\% accuracy.
This performance benefit is consistent across social science questions, questions with text or no contexts, and higher-grade questions.
More importantly, we noticed that this model reaches a comparable performance with the LLaVA model, which is trained with synthetic data and twice the size of the training data than what the former model has seen, and in some cases has additional support from GPT-4 during inference.

Second, we noticed that \textit{LLaMA-SciTune-ScienceQA-7B (CTOM)} model performs better than \textit{LLaMA-SciTune-ScienceQA-7B (C)} model pretrained only with captions.
For example, CTOM variant (86.11) slightly outperforms C variant (85.11) on average performance and across many other sub-groupings.
This suggests the importance of interleaved multimodal data in the scientific concept alignment stage which lifts the downstream task performance over the model tuned only with figure-caption data.

Finally, we noticed a significant performance advantage of the models trained with larger language decoder model (13B) compared to the relatively smaller model (7B).
For example, the \textit{LLaMA-SciTune-ScienceQA-13B (CTOM)} model has nearly 5\% performance advantage over the 7B model variant.
This advantage is 5x bigger than what reported by the LLaVA model when scaled from 7B to 13B~\citep{liu2023visual}.
While this observation suggests that the larger language decoder model helps to improve the multimodal reasoning performance, we believe it could lead to huge performance benefit with even larger models (LLaMA-65B) when trained with highly-curated scientific multimodal instruction tuning datasets.

\paragraph{\textbf{Explanation Performance Analysis}}
In addition to generating the specific answers to the questions asked, \textit{LLaMA-SciTune-ScienceQA} models also generate a corresponding lecture and explanation for the answers.
Please see Figures ~\ref{fig:scitune_example_multimodal} and ~\ref{fig:scitune_example_unimodal} and in the Appendix for several examples of generated lectures and explanations.
In order to better understand the behavior of generated solution, we manually investigate a few random test examples. Specifically, we picked $50$ samples from both the correct and incorrect predictions. 
We observe that even the correct samples contain a certain amount of incorrect solutions, i.e., around $8\%$ in C and $2\%$ in CTOM version of the 7B models. These results indicate that solution may not always benefit the final answer, and the model is robust to some extent, i.e., it can predict the correct answer even with incorrect rationales. The incorrect solutions are further divided into two major categories, namely commonsense that requires commonsense knowledge such as factual information and counting numbers in the images, and the logical mistakes which shows contradictions in the reasoning. In our experiment, commonsense mistakes are dominant compared to logical, which aligns with previous work~\citep{zhang2023multimodal}. 
Furthermore, there are cases where solutions are correct in an absolute sense but their final answers are wrong. 
We also noticed that solutions generated by the CTOM version of the model are more accurate compared to the C version of the model, further emphasizing the importance of multi-modal training with additional scientific modalities. 
There are certain task categories where our model performs extremely well compared to baselines. 
In our manual analysis, we found the model is very good with numerical questions, including temperatures and distances, and can answer all topological/map related questions such as "which ocean is highlighted" in the image.  

\ignore{
\begin{table}[htbp]
\caption{Explanation analysis of LLaMA-SciTune 7B models}
\label{tab:error_analysis}
\scalebox{0.85}{
\begin{tabular}{|c|c|c|c|}
\hline
\multirow{2}{*}{Answers}           & \multirow{2}{*}{Explanation Type} & \multicolumn{2}{c|}{Percentage} \\ \cline{3-4}
                           &                            & \multicolumn{1}{c|}{C}       & CTOM      \\ \hline \hline
\multirow{3}{*}{Correct}   & Correct                    & 
\multicolumn{1}{c|}{92}        &    98       \\ \cline{2-4} 
                           & Incorrect-Logical          &  \multicolumn{1}{c|}{4}        &     0      \\ \cline{2-4} 
                           & Incorrect-Common sense     & \multicolumn{1}{c|}{4}        &      2     \\ \hline
\multirow{3}{*}{Incorrect} & Correct                    & \multicolumn{1}{c|}{4}        &     2      \\ \cline{2-4} 
                           & Incorrect-Logical          & \multicolumn{1}{c|}{2}        &     {2}      \\ \cline{2-4} 
                           & Incorrect-Common sense     & \multicolumn{1}{c|}{94}        &    {96}       \\ \hline
\end{tabular}
}
\end{table}
}


While we observe high performance in aggregate, it is also important to determine whether this performance persists in cases with minimal training examples.
We evaluate the performance of the model for questions whose accompanying lectures are only observed a few times in the training data.
For these few-shot examples, the model will be less likely to have the exact lecture memorized and ready to use in its generation of the answer, which could lead to lower performance.

We show the model performance on questions for which the lectures were viewed in 5, 10, 25, and 50 times during training, in Table \ref{tab:few_shot_performance}.
The model performance drops substantially for questions with only 5 or fewer lectures in the training data but quickly recovers after the lecture is viewed at least 10 times.
This suggests that the \textit{LLaMA-SciTune-ScienceQA} model doesn't require substantial exposure to a particular type of knowledge to achieve adequate performance.
Furthermore, this performance drop is worse for the 7B model as compared to the 13B model, which means that the 13B model is able to learn more quickly from fewer examples or may have more knowledge ``baked in'' from pretraining that can be leveraged for few-shot examples.
Future extensions of the model to other datasets should test performance on completely unseen data, e.g. a more standard VQA dataset not used during training, to determine whether the model is similarly robust in other domains.

\begin{table}[htbp]
\centering
\caption{Few-shot performance analysis. We report the number of times lectures seen during the training in frequency, and the number of test questions with the lecture.}
\label{tab:few_shot_performance}
\scalebox{0.9}{
\begin{tabular}{|p{1.5cm}| p{1.5cm}| p{1.5cm}| p{1.5cm}|}
\hline
Frequency & \#Questions & Accuracy (7B) & Accuracy (13B) \\ \hline
5                                                      & 36                       & 75.00 & 83.33 \\
10                                                     & 125                      & 81.60 & 85.60 \\
25                                                     & 412                      & 80.34 & 85.92 \\
50                                                     & 1140                     & 81.05 & 86.14 \\ \hline
\end{tabular}
}
\end{table}


\paragraph{\textbf{Chain of Thought Reasoning Performance}}
Outside of the coarse-grained accuracy metric (did the model get the answer right?), we also need to determine whether the model's overall process of reasoning was correct (did the model accurately explain the reasoning that supports the answer?).
We investigate the accuracy of the generated text, outside of the answer alone,
assessing if the model is able to accurately recover the lecture and the solution that it was trained to generate and to help its reasoning toward the final answer.
We report the BLEU and ROUGE scores over all the generated text, separated into the lecture and solution components and compared with the corresponding ground-truth data, e.g. compare the generated lecture component with the ground-truth lecture.

The aggregate results for the generation metrics are shown in Table \ref{tab:evaluation_text_generation}.
When considering all the questions, the model generates the solution text with higher accuracy than the lecture text.
However, in cases where the model answers incorrectly, the trend reverses and the model has a higher accuracy in generating the lecture text as compared to the solution text. 
Therefore, the model may be failing to answer these questions due to a failure to reason in the ``solution stage'' of its generation.
Furthermore, for the 13B model we see that the lecture generation performance is higher for incorrect answers than correct answers (ROUGE score of 0.924 for incorrect vs. 0.861 for correct).
This could indicate overfitting, where the model ``memorizes'' lectures that apply to the problem but fails to apply the lectures to the actual solution.

This problem is apparent with an example question about object properties, where the model must determine the property shared by an icicle, a fish bowl, a glass, and a tea cup.
The model correctly generates the lecture about object properties required to reason through the problem (``An object has different properties. A property of an object can tell you how it looks, feels, tastes, or smells.'').
However, in the solution stage the model incorrectly reasons that all the objects were transparent instead of fragile, based on a failure to infer the properties of the objects from the image (``You can see clearly through a transparent object. All four objects are transparent.'').

Incorrect reasoning can be attributed to two factors, i.e., linguistic and visual features. 
In a manual analysis of $100$ test samples, we found that linguistic features are a weakness for mainly two use cases, namely retrieving commonsense facts (e.g. characteristics of bird song) and semantic understanding of words in terms of figure of speech and relative position of words in the dictionary. 
In contrast, visual features appear to be strong in use-cases such as identifying geographic areas but it lags in counting numbers in images and retrieving properties of objects such as color, texture and states.

These observations suggest that to improve model training, we need a wide variety of human-curated instructions, especially datasets that include both text and visuals explained by humans.
Such diverse explanations would help the model understand various scenarios that require different types of reasoning.


\begin{table}[htbp]
\caption{Evaluation of generated lectures and solutions.}
\label{tab:evaluation_text_generation}
\centering
\scalebox{0.9}{
\begin{tabular}{|p{1cm} r r | r r|}
\hline
\multicolumn{1}{|l}{} & \multicolumn{2}{c}{7B Model} & \multicolumn{2}{|c|}{13B Model} \\
~            & BLEU          & ROUGE        & BLEU          & ROUGE         \\ \hline
\multicolumn{3}{|l}{All answers}  & \multicolumn{2}{|l|}{}                                                   \\

Lecture              & 0.763       & 0.778     & 0.854      & 0.868      \\
Solution             & 0.791      & 0.838     & 0.872      & 0.921      \\ \hline
\multicolumn{3}{|l}{Correct answers}    & \multicolumn{2}{|l|}{}                                                     \\
Lecture              & 0.765      & 0.780     & 0.847      & 0.861      \\
Solution             & 0.829      & 0.873     & 0.893      & 0.937      \\ \hline
\multicolumn{3}{|l}{Incorrect answers}     & \multicolumn{2}{|l|}{}                                                  \\
Lecture              & 0.751      & 0.767     & 0.909       & 0.924      \\
Solution             & 0.565      & 0.631     & 0.694      & 0.778 \\ \hline
\end{tabular}
}
\end{table}

%% file: PNNL-ST-SciTune/sections/scienceqa_baseline_results.tex
\begin{table*}[!t]
\centering
\caption{Results (accuracy \%) on ScienceQA dataset. Question classes: NAT = natural science, SOC = social science, LAN = language science, TXT = text context, IMG = image context, NO = no context, G1-6 = grades 1-6, G7-12 = grades 7-12. We present two variants, \textit{LLaMA-SciTune-ScienceQA (C)} and \textit{LLaMA-SciTune-ScinceQA (CTOM)}. Acronyms inside the parenthesis represent the text inputs used in the SciTune instruction template. E.g., \textbf{C}aption, Figure \textbf{T}ype, \textbf{O}CR, and Figure \textbf{M}entions. We use the notation $\spadesuit$ to denote the models finetuned with GPT-3.5/4 synthetic instructions, or use GPT-3.5/4 for any support during the inference time. We bold the accuracy values that are greater than what humans achieved. 
}
\label{fig:sqa_results}
\scalebox{0.85}{
\begin{tabular}{|l|r|l|l|l|l|l|l|l|l|l|}
\hline
\textbf{Method}     & \textbf{\#Params} & \textbf{Avg} & \textbf{NAT} & \textbf{SOC} & \textbf{LAN} & \textbf{TXT} & \textbf{IMG} & \textbf{NO} & \textbf{G1-6} & \textbf{G7-12} \\ \hline \hline
Random Chance       & -                & 39.83        & 40.28        & 46.13        & 29.25        & 47.45        & 40.08        & 33.66       & 39.35         & 40.67          \\ \hline
Human Average       & -                & 88.40        & 90.23        & 84.97        & 87.48        & 89.60        & 87.50        & 88.10       & 91.59         & 82.42          \\ \hline
UnifiedQA           & 223M             & 70.12        & 68.16        & 69.18        & 74.91        & 63.78        & 61.38        & 77.84       & 72.98         & 65.00          \\ \hline
UnifiedQA (CoT)     & 223M             & 74.11        & 71.00        & 76.04        & 78.91        & 66.42        & 66.53        & 81.81       & 77.06         & 68.82          \\ \hline
$\spadesuit$ GPT-3 (Zero Shot)   & 175B               & 74.04        & 75.04        & 66.59        & 78.00        & 74.24        & 65.74        & 79.58       & 76.36         & 69.87          \\ \hline
$\spadesuit$ GPT-3 (CoT) (ALE)   & 175B               & 75.17        & 75.44        & 70.87        & 78.09        & 74.68        & 67.43        & 79.93       & 78.23         & 69.68          \\ \hline
$\spadesuit$ ChatGPT CoT         & 175B+               & 78.31        & 78.82        & 70.98        & 83.18        & 77.37        & 67.92        & 86.13       & 80.72         & 74.03          \\ \hline
$\spadesuit$ GPT-4 CoT           & 1T+               & 83.99        & 85.48        & 72.44        & \textbf{90.27}        & 82.65        & 71.49        & \textbf{92.89}       & 86.66         & 79.04          \\ \hline
Multimodal-CoT      & 223M             & 84.91        & 87.52        & 77.17        & 85.82        & 87.88        & 82.90        & 86.83       & 84.65         & \textbf{85.37}          \\ \hline
Multimodal-CoT      & 770M             & \textbf{91.68}        & \textbf{95.91}        & 82.00        & \textbf{90.82}        & \textbf{95.26}        & \textbf{88.80}        & \textbf{92.89}       & \textbf{92.44}         & \textbf{90.31}          \\ \hline
$\spadesuit$ LLaMA-Adapter       & 13B           & 85.19        & 84.37        & \textbf{88.30}        & 84.36        & 83.72        & 80.32        & 86.90       & 85.83         & \textbf{84.05}          \\ \hline
$\spadesuit$ LLaVa              & 13B              & \textbf{90.92}        & \textbf{90.36}        & \textbf{95.95}        & 88.00        & 89.49        & \textbf{88.00}        & \textbf{90.66}       & 90.93         & \textbf{90.90}          \\ \hline
$\spadesuit$ LLaVa + GPT-4 (judge)       & 13B              & \textbf{92.53}        & \textbf{91.56}        & \textbf{96.74}        & \textbf{91.09}        & \textbf{90.62}        & \textbf{88.99}        & \textbf{93.52}       & \textbf{92.73}         & \textbf{92.16}          \\ \hline
$\spadesuit$ Chameleon (ChatGPT) & 175B+               & 79.93        & 81.62        & 70.64        & 84.00        & 79.77        & 70.80        & 86.62       & 81.86         & 76.53          \\ \hline
$\spadesuit$ Chameleon (GPT-4)   & 1T+               & 86.54        & 89.83        & 74.13        & \textbf{89.82}        & 88.27        & 77.64        & \textbf{92.13}       & 88.03         & \textbf{83.72}          \\ \hline
SciTune-ScienceQA (C)  & 7B & 85.61        &  84.36       &  \textbf{92.23}   &   82.81    &  89.56     &   81.26     &   \textbf{88.29}   &  81.28    &   \textbf{86.03}      \\ \hline
SciTune-ScienceQA (CTOM)     &  7B & 86.11        &  84.50       &  \textbf{94.15}   &    82.91   &   88.35    &   83.64     &   \textbf{88.74}   &    85.05  & \textbf{85.60}         \\ \hline
SciTune-ScienceQA (CTOM)    &  13B & \textbf{90.03}        &   89.30    &  \textbf{95.61} & 87.00   & \textbf{93.08}    & 86.67      &   \textbf{91.75}   & 84.37  &   \textbf{91.30}     \\ \hline
\end{tabular}
}
\end{table*}

%% file: PNNL-ST-SciTune/sections/related.tex
\section{Related Work}
\citet{zhang2023llama} proposed LLaMA-Adapter to guide the LLaMA model to follow multimodal instructions.
Specifically, they proposed a zero-init attention with gating as a Parameter-Efficient Fine-Tuning (PEFT) technique to prepend learnable multimodal adaptation prompts to the input text tokens at higher transformer layers in the LLaMA model.
The same authors proposed LLaMA-Adapter-V2~\citep{gao2023llamaadapter} that distributes the learnable parameters across all layers in the LLaMA model to improve performance in multimodal reasoning.
MiniGPT-4~\citep{zhu2023minigpt} combined the frozen LLM (Vicuna) and a vision encoder with a single projection layer and finetuned with a highly-curated visual conversation dataset.
More recently,~\citet{liu2023visual} introduce \textit{visual instruction tuning} to develop general-purpose visual assistant (LLaVA) that follows multimodal instructions.
They present several data reformation techniques to construct multimodal instruction-following data from the standard image-text pairs.
For example, the LLaVA model was trained with 595K image-text pairs filtered from the CC3M dataset~\citep{sharma2018conceptual}, and 158K unique language-image instruction-following data generated from ChatGPT/GPT-4~\citep{liu2023visual}.
This multimodal instruction set includes image-based \textit{conversations} and \textit{detailed descriptions} and \textit{complex reasoning} questions.
LLaVA~\citep{liu2023visual} reaches the best performance in the ScienceQA benchmark with support from GPT-4 that acts as a judge to evaluate the generated answers.
LLaVAR~\citep{zhang2023llavar} extends the LLaVA for text-rich images by training with additional 422K image-OCR and 16K conversations generated from GPT-4.

%% file: PNNL-ST-SciTune/sections/conclusion.tex
\section{Conclusion}
In this work, we present scientific multimodal instruction tuning to align LLMs with scientific concepts and goals.
To this end, we use human-generated multimodal instructions curated from visual signals (e.g., plots, charts, equations), and textual signals (e.g., captions, optical character recognition (OCR) and paragraph mentions) found within scientific publications.
We train several models built on top of LLaMA language decoder model and CLIP vision encoder model and test the models on science-focused multimodal downstream tasks.
In evaluation, we show that the resulting \textit{LLaMA-SciTune-SciCap} models can perform better on classifying scientific visuals and generating figure captions compared with SOTA vision-to-language models.
Furthermore, the \textit{LLaMA-SciTune-ScienceQA} model surpasses the human performance in ScienceQA, the standard multimodal science-focused reasoning QA benchmark.

Our results suggest human-curated scientific multimodal data remains highly valuable despite the advancements in synthetic data generation techniques.
While it is fast and easy to generate large volume of synthetic training data with closed-source models such as GPT-4, they may contain inaccuracies or biases due to lack of expert review.
In contrast, although existing human-generated scientific multimodal datasets are comparatively smaller, 
they provide reliable ground truth for tuning LLMs, which leads to better generalization and performance in downstream science applications.


%% file: PNNL-ST-SciTune/sections/ack.tex
\section*{Acknowledgements}
This work was supported by the NNSA Office of Defense Nuclear Nonproliferation Research and Development, U.S. Department of Energy, and Pacific Northwest National Laboratory, which is operated by Battelle Memorial Institute for the U.S. Department of Energy under Contract DE-AC05–76RLO1830. This article has been cleared by PNNL for public release as PNNL-SA-186641.

%% file: PNNL-ST-SciTune/sections/appendix.tex
\appendix

\section{SciTune Multimodal Instructions}
Table~\ref{tab:scitune_example} shows two SciTune instruction examples used in the scientific concept alignment training stage.
Table~\ref{tab:llava_instructions} presents the LLaVA questions that used to sample the $s_I$ instructions.

\begin{table*}[htbp]
    \centering
    \caption{SciTune Multimodal Instruction Examples. We distinguish the system message \textcolor{gray}{$s_D$}, natural language instruction \textcolor{red}{$s_I$}, scientific figure type \textcolor{green}{$s_t$}, caption \textcolor{blue}{$s_c$}, OCR \textcolor{orange}{$s_o$} and paragraph mentions \textcolor{purple}{$s_m$}. The list of instructions used to sample $s_{I}$ are presented at Table~\ref{tab:llava_instructions}.}
    \label{tab:scitune_example}
    \begin{tabular}{l}
         \textcolor{gray}{A chat between a curious human and an artificial intelligence assistant. The assistant gives helpful,} \\ \textcolor{gray}{detailed, and polite answers to the human’s questions.} \\
         \textcolor{gray}{Human:} \textcolor{red}{Give an elaborate explanation of the image you see.}\\
         \includegraphics[scale=0.2]{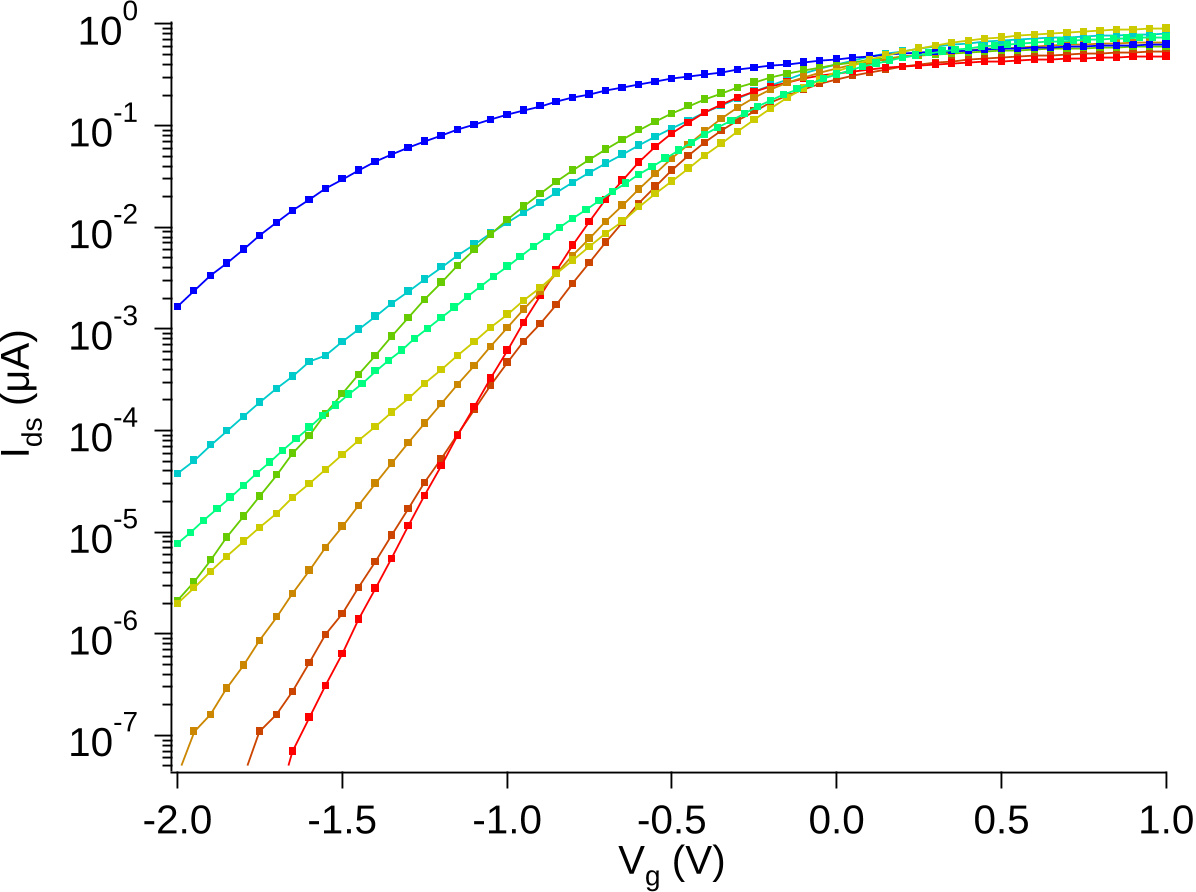} \\
         \textcolor{gray}{Assistant:} \textcolor{green}{Graph Plot} \textcolor{blue}{Room temperature characteristics (drain-source current versus gate voltage) for a set} \\
         \textcolor{blue}{of nominally identical samples from the same wafer. Sample to sample variations
         are very important,}  \\ 
         \textcolor{blue}{especially below the threshold voltage (-0.5 V for the best device, in red). In the most extreme case (blue curve)}  \\ 
         \textcolor{blue}{the many decades of excess current compared to the best device are attributed to direct transport through} \\ 
         \textcolor{blue}{individual arsenic dopants which have diffused into the channel after doping and annealing steps.} \\
         \textcolor{blue}{The source-drain voltage is 10mV for these data.} \\
         \textcolor{purple}{FIG. 6: figure S1: Room temperature characteristics (drain-source current versus gate voltage) for} \\ 
         \textcolor{purple}{a set of nominally identical samples from the same wafer.} \\ \\
         \textcolor{gray}{A chat between a curious human and an artificial intelligence assistant. The assistant gives helpful,} \\ \textcolor{gray}{detailed, and polite answers to the human’s questions.} \\
         \textcolor{gray}{Human:} \textcolor{red}{Describe the following image in detail.}\\
         \includegraphics[scale=0.5]{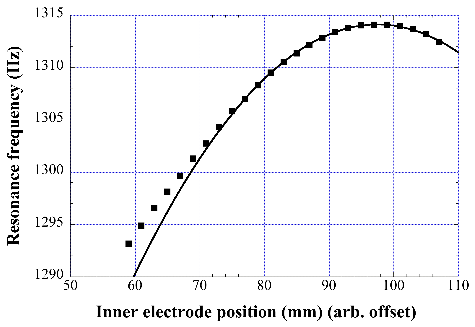} \\
         \textcolor{gray}{Assistant:} \textcolor{green}{Graph Plot} \textcolor{blue}{Resonance frequency of the (1, 0) mode of the r1 = 4 cm cavity filled with 1 atm} \\ 
         \textcolor{blue}{of N2 as a function of the vertical position of the central electrode.} \\
         \textcolor{orange}{Inner electrode position, (arb , offset)} \\
         \textcolor{purple}{In Fig. 10 we have an example of this behavior: the resonance frequency shows a maximum} \\
         \textcolor{purple}{which corresponds to a minimum in the eccentricity of the electrode [39].} \\
    \end{tabular}
\end{table*}

\begin{table*}[htbp]
    \caption{LLaVA Instructions for detailed image descriptions}
    \label{tab:llava_instructions}
    \centering
    \begin{tabular}{|l|}
    \hline
                      "Describe the following image in detail." \\
    "Provide a detailed description of the given image." \\
    "Give an elaborate explanation of the image you see." \\
    "Share a comprehensive rundown of the presented image."  \\
    "Offer a thorough analysis of the image."  \\
    "Explain the various aspects of the image before you."  \\
    "Clarify the contents of the displayed image with great detail."  \\
    "Characterize the image using a well-detailed description."  \\
    "Break down the elements of the image in a detailed manner."  \\
    "Walk through the important details of the image."  \\
    "Portray the image with a rich, descriptive narrative."  \\
    "Narrate the contents of the image with precision."  \\
    "Analyze the image in a comprehensive and detailed manner."  \\
    "Illustrate the image through a descriptive explanation."  \\
    "Examine the image closely and share its details."  \\
    "Write an exhaustive depiction of the given image."  \\\hline
    \end{tabular}
\end{table*}

\section{Training Details}
We use the LLaVA codebase~\citep{liu2023visual} for multimodal adapter training with SciTune instructions.
We train the model for 1 epoch with 128 batch size with a 0.002 learning rate and 2048 context length.
LLaMA-SciTune 7B and 13B model variants took 6.5 and 11.2 hours to train, respectively with 8 x A100 GPUs.
We use Pytorch Fully Sharded Data Parallel (FSDP) to recursively wrap the language models decoder layers in the task-specific instruction finetuning stage.
We finetuned the models for 12 epochs with the ScienceQA training example to make a fair comparison with LLaVA.

\section{LLaMA-1 and LLaMA-2 Base Model Comparison}
\begin{figure}[htbp]
    \centering
    \includegraphics[scale=0.5]{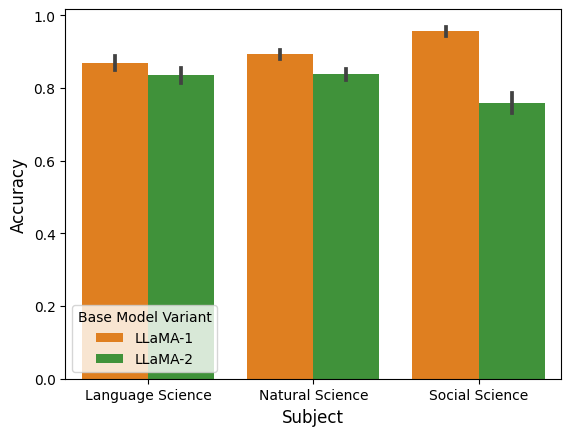}
    \caption{ScienceQA Performance of the LLaMA-SciTune models with LLaMA-1 and LLaMA-2 as the base language decoder models}
    \label{fig:scienceqa_llama}
\end{figure}

We also test our methodology with LLaMA-2~\citep{touvron2023llama2} as the base language decoder model.
LLaMA-2 was reported to have superior performance compared to LLaMA-1 with additional pretraining corpus (2x tokens), larger context length (2x), and adopted grouped-query attention.
In addition, there were additional steps taken to improve the safety of LLaMA-2 models.
We do not use the LLaMA-2 model variants optimized for chat and dialogue use cases to make a fair comparison with LLaMA-1 model.
In this experiment, we repeat the entire training (CTOM) pipeline (as shown in Figure~\ref{fig:scitune_framework}) with the LLaMA-2 (13B) model.
Figure~\ref{fig:scienceqa_llama} shows a comparison of ScienceQA performance with LLaMA-1 and LLaMA-2 base language models.
Despite the reported performance improvements in the LLaMA-2 model over the LLaMA-1, we do not observe any performance advantage of the former model in the ScienceQA benchmark.
This may be due to the effect of larger pretraining data or a different dataset mix used to pretrain LLaMA-2.
For example, LLaMA-2 authors reported an increase in toxicity of the new models with more than 7B parameters~\citep{touvron2023llama2}.
It remains as a future work to reason this performance difference with more empirical results.
We use the LLaMA-1 as the base language decoder model in the rest of the experiments unless explicitly mentioned.


\section{Visual Grounded Task Performance}
Table~\ref{tab:gen_caps} shows a few generated captions for the SciCap images used to test the model performance on visual grounded tasks.
We report the gold-standard captions as they appeared in the arXiv articles used to collect SciCap dataset, and the captions generated from the BLIP and LLaMA-SciTune (13B, CTOM) models for the comparisons.

\begin{table*}[htbp]
    \caption{A Sample of Generated Captions. We highlight the gold standard caption in red, and generated captions from the BLIP~\citep{li2022blip} model in gray. LLaMA-SciTune model first generates the figure types followed with the captions colored in blue.}
    \label{tab:gen_caps}
    \centering
    \scalebox{0.90}{
    \begin{tabular}{p{3cm}p{12cm}}
         \multirow{4}{*}{\includegraphics[scale=0.14]{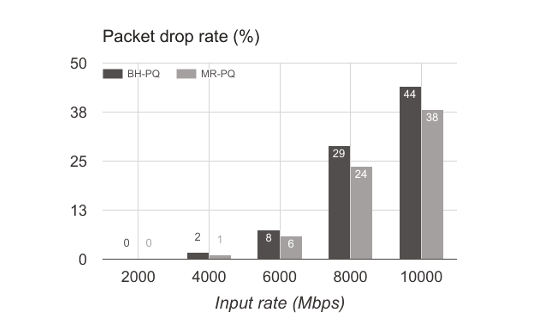}} & \textcolor{red}{Packet drop rate} \\ 
         & \textcolor{gray}{a chart of a bar chart with a number of different items} \\
         & \textcolor{blue}{\textbf{Bar Chart} Packet drop rate for each method.} \\ 
         & \\ \\
         \multirow{4}{*}{\includegraphics[scale=0.14,]{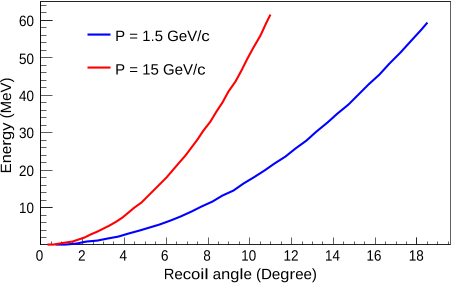}} & \textcolor{red}{ The kinetic energy of the recoil protons as a function of the recoil angle at beam momenta P=1.5 and 15 GeV/c, blue and red, respectively.} \\ 
         & \textcolor{gray}{  a plot of a curve with a blue line and a red line.} \\
         & \textcolor{blue}{\textbf{Graph Plot} The angular distribution of the electron recoil spectrum in the 1.5 GeV/c and 15 GeV/c electron beams.} \\ 
         & \\
         \multirow{4}{*}{\includegraphics[scale=0.11]{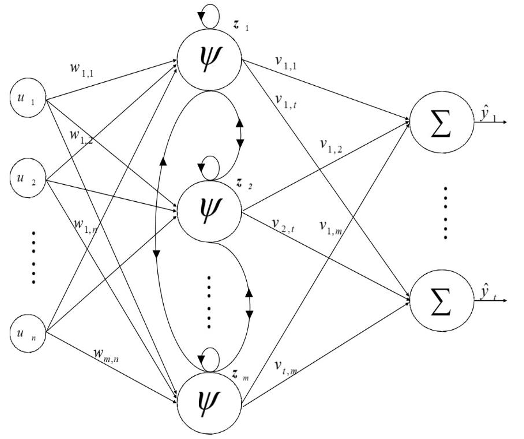}} & \textcolor{red}{ Artificial neural network structure.} \\ 
         & \textcolor{gray}{  a diagram of a network with several different paths.} \\
         & \textcolor{blue}{\textbf{Node Diagram} The generative neural network.} \\ 
         & \\ \\
        \multirow{4}{*}{\includegraphics[scale=0.14]{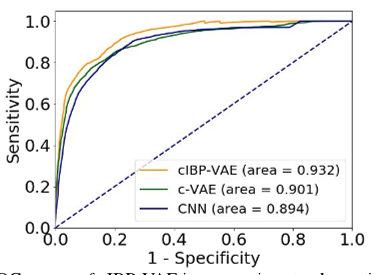}} & \textcolor{red}{ ROC curves of cIBP-VAE in comparison to alternative models on the clinical ECG data set.} \\ 
         & \textcolor{gray}{ a plot of the average and average time of a cell phone.}\\
         & \textcolor{blue}{\textbf{Graph Plot} ROC curves of c-VAE, CNN, and c-VAE+CNN on the cerebellar atrophy dataset.} \\ 
         & \\
         \multirow{4}{*}{\includegraphics[scale=0.11]{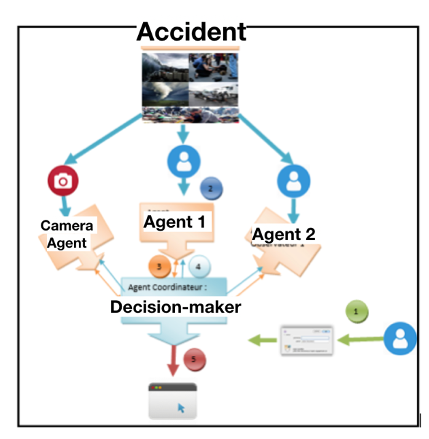}} & \textcolor{red}{ Functional architecture of the developed prototype.} \\
         & \textcolor{gray}{ a diagram of a camera and a person on a phone.} \\
         & \textcolor{blue}{\textbf{Node Diagram} An overview of the system architecture of the proposed method.}\\ 
         & \\ \\
         \multirow{4}{*}{\includegraphics[scale=0.07]{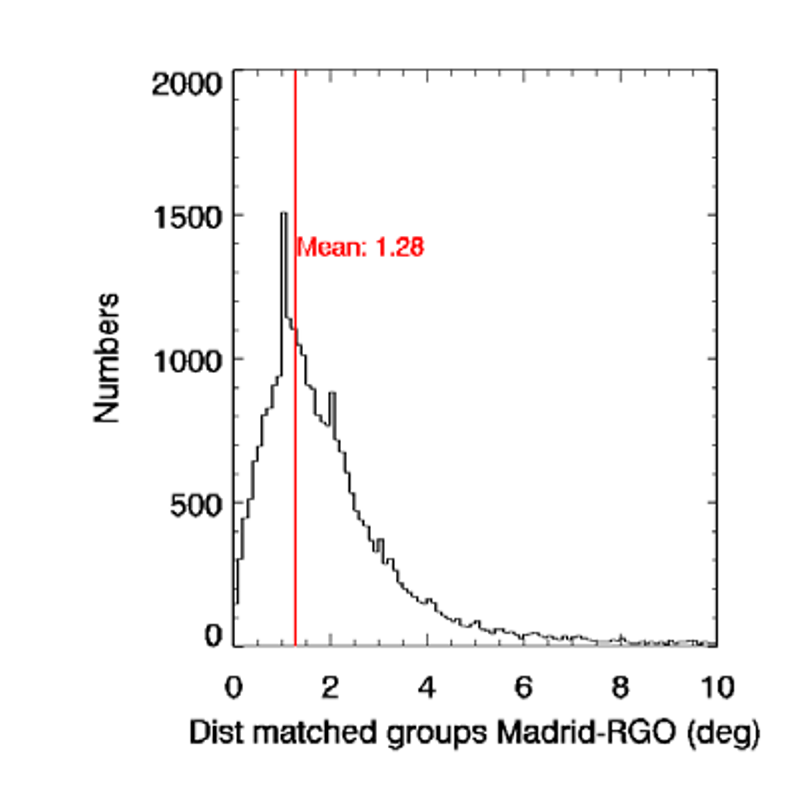}} & \textcolor{red}{ Distance between matched groups in Madrid and RGO catalogs (bins of 0.1 degrees). The red line represents the mean value.} \\ 
         & \textcolor{gray}{ a plot of a line of data with a red line and a white line.} \\
         & \textcolor{blue}{\textbf{Graph Plot} Distance correlation between groups matched by Madrid RGO.} \\ 
         & \\ \\
         \multirow{4}{*}{\includegraphics[scale=0.07]{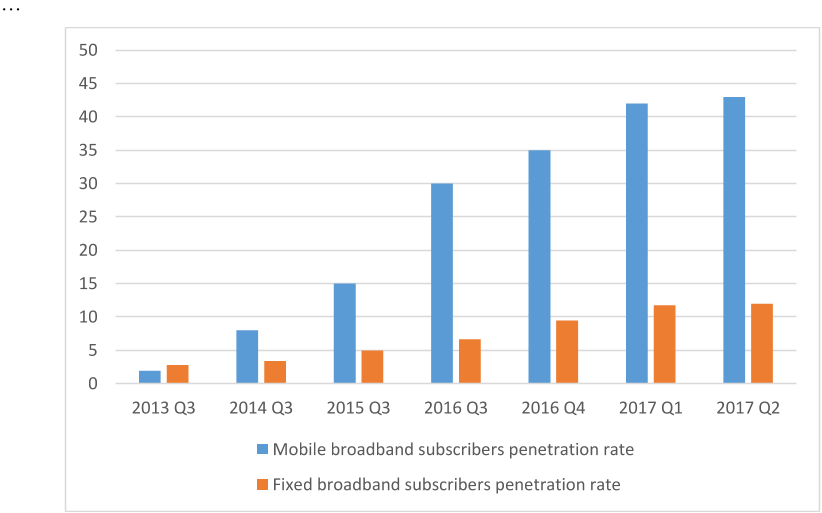}} & \textcolor{red}{ Comparison of penetration rate of mobile broadband subscribers with that of fixed broadband subscribers.} \\ 
         & \textcolor{gray}{ a chart of the number of people who are using the internet.} \\
         & \textcolor{blue}{\textbf{Bar Chart} The average rate of mobile broadband subcribers and fixed broadband subscribers for each quarter (in thousands).} \\ 
         & \\
         \multirow{4}{*}{\includegraphics[scale=0.16]{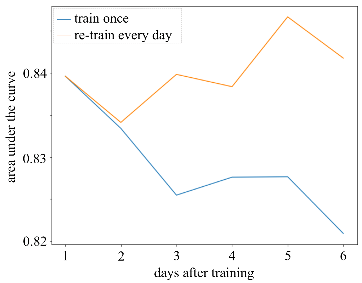}} & \textcolor{red}{ Model staleness of the one-off trained model vs. the model retrained every day.} \\ 
         & \textcolor{gray}{ a diagram of a graph with a line graph and a line graph.} \\
         & \textcolor{blue}{ \textbf{Graph Plot} Prediction accuracy of model trained once and fine-tuned every day.} \\ 
         & \\ \\
         \multirow{4}{*}{\includegraphics[scale=0.15]{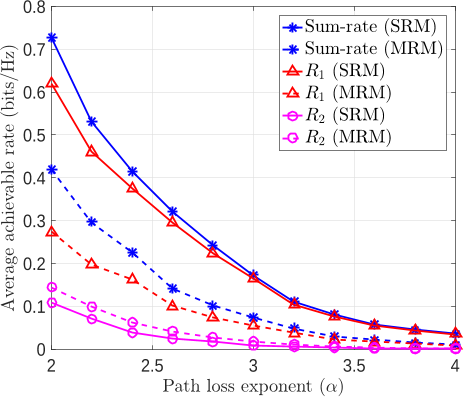}} & \textcolor{red}{ Comparison of the effect of the path loss exponent $\alpha$ on rates achieved by both transmitters, M = 4.} \\ 
         & \textcolor{gray}{ a plot of a line graph with a blue line and red line.} \\
         & \textcolor{blue}{ \textbf{Graph Plot} The sum-rate and sum-rate of RRM-RRM with respect to the path loss exponent $\gamma$ for the two cases: $\gamma=2$ and $\gamma=3$.} 
         \\ 
         & \\ \\
         \multirow{4}{*}{\includegraphics[scale=0.06]{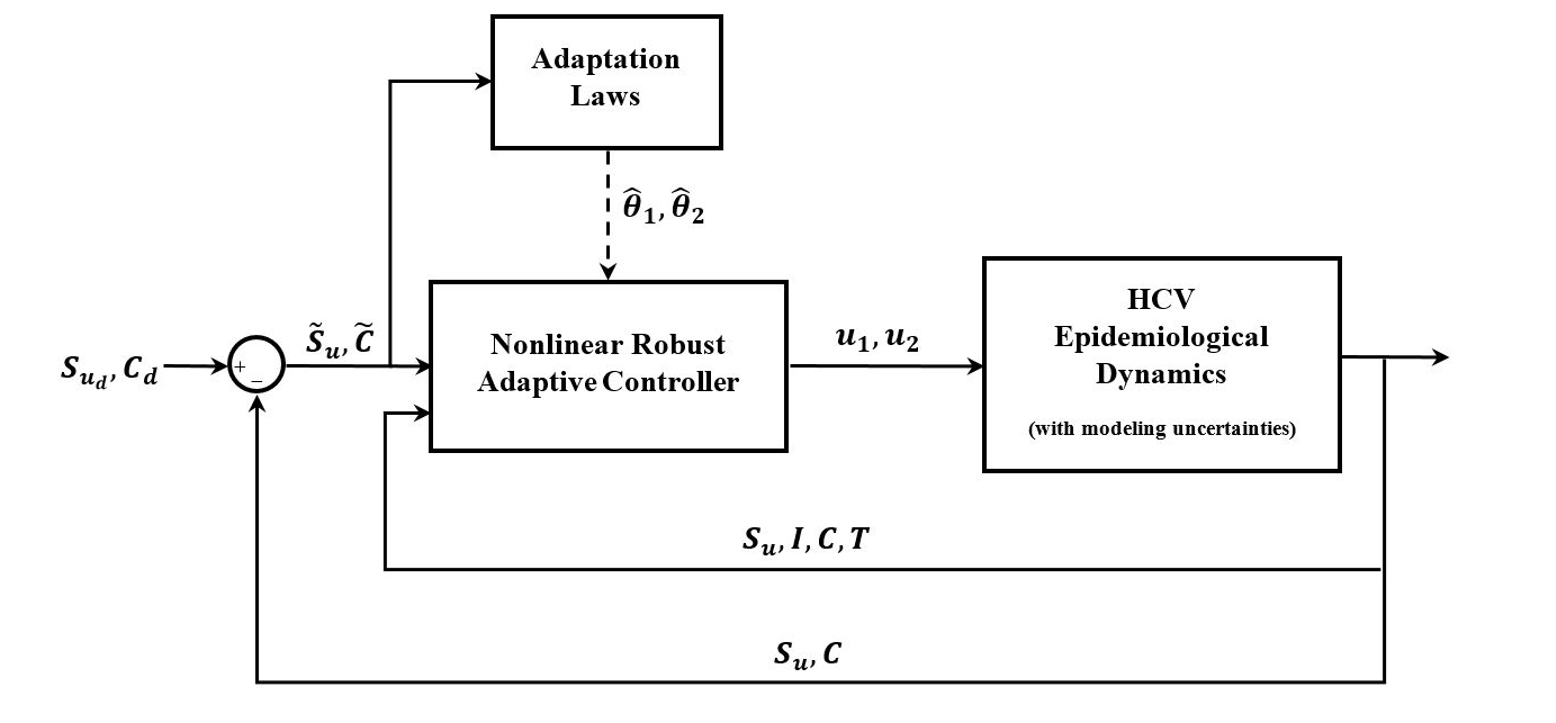}} & \textcolor{red}{ Conceptual diagram of nonlinear adaptive method developed to control the HCV epidemic in the existence of uncertainties on parameters of the model.} \\ 
         & \textcolor{gray}{ a diagram of a block diagram of a nuclear system.} \\
         & \textcolor{blue}{\textbf{Node Diagram} Block diagram of the proposed non-linear SIR epidemic model with adaptive controllers.}
         \\ 
         & \\
    \end{tabular}
    }
\end{table*}

\section{ScienceQA Chain of Thought Reasoning Examples}
\begin{figure*}[!t]
\centering
\begin{subfigure}[b]{0.95\textwidth}
   \includegraphics[width=1\linewidth]{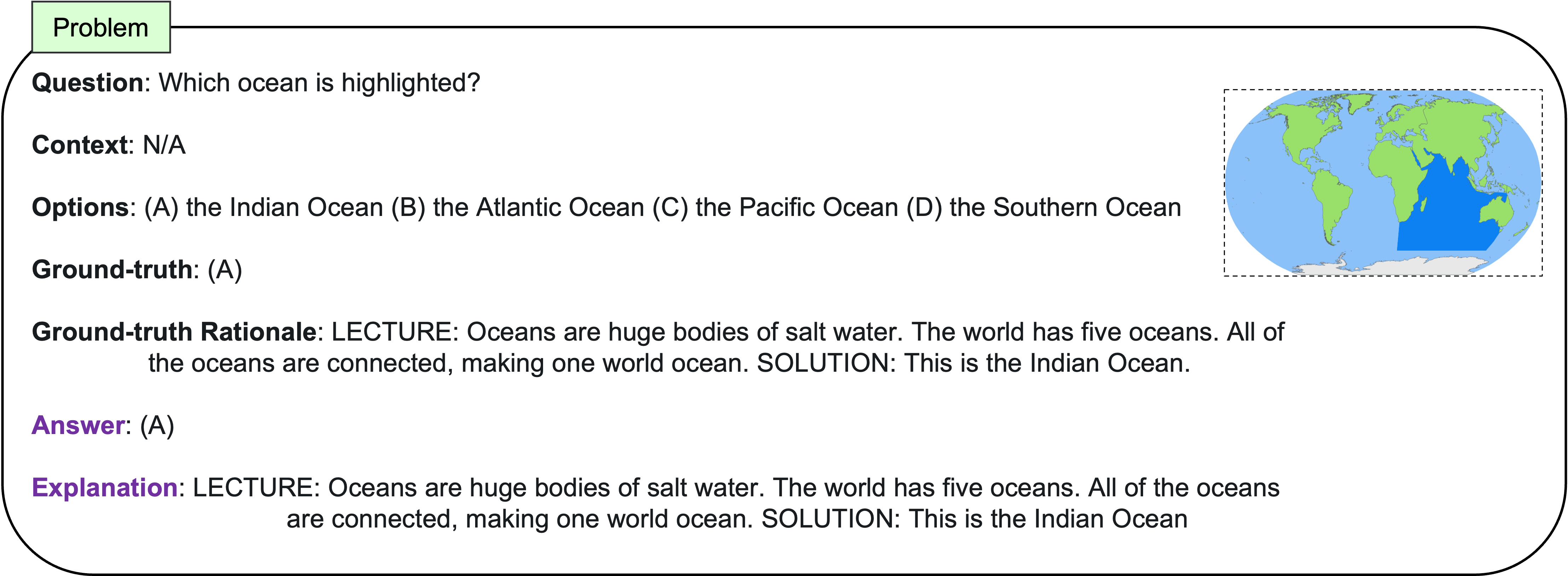}
   \caption{An example with right answer and right explanation}
   \label{fig:scitune_example_1} 
\end{subfigure}
\begin{subfigure}[b]{0.95\textwidth}
   \includegraphics[width=1\linewidth]{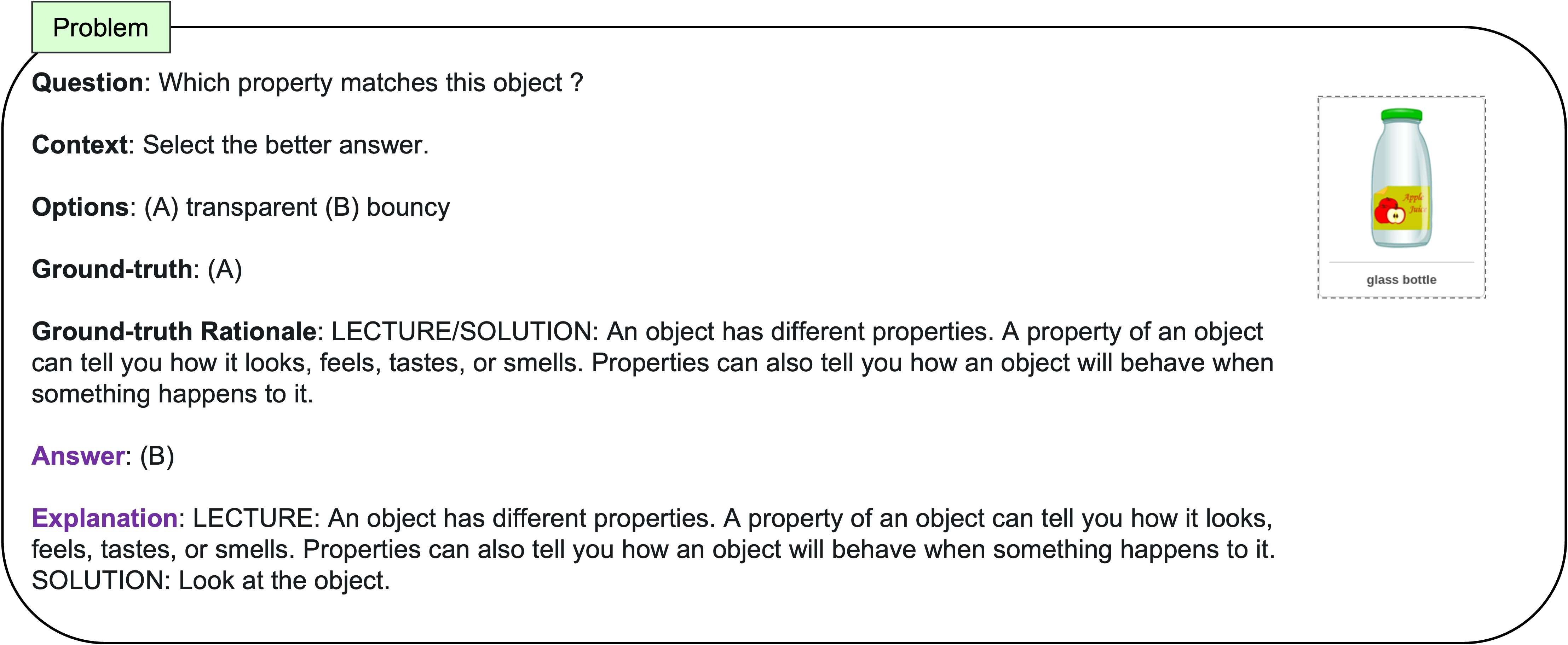}
   \caption{An example with incorrect answer and incorrect explanation}
   \label{fig:scitune_example_2}
\end{subfigure}
\caption{Two Multimodal QA examples with answer and explanation generated by LLaMA-SciTune}
\label{fig:scitune_example_multimodal}
\end{figure*}

\begin{figure*}[!t]
\centering
\begin{subfigure}[b]{0.95\textwidth}
   \includegraphics[width=1\linewidth]{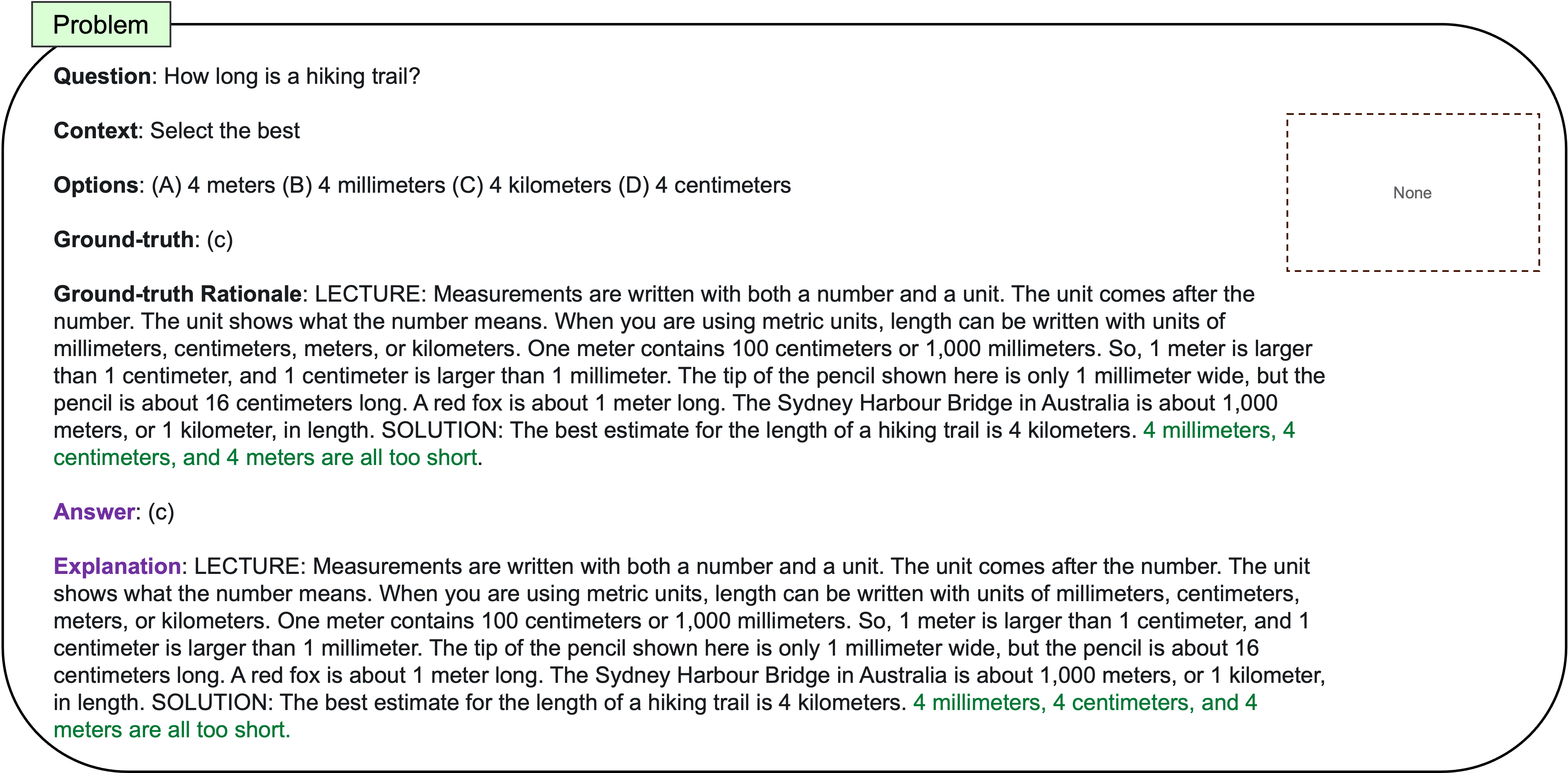}
   \caption{An example with right answer and right explanation}
   \label{fig:scitune_example_3} 
\end{subfigure}
\begin{subfigure}[b]{0.95\textwidth}
   \includegraphics[width=1\linewidth]{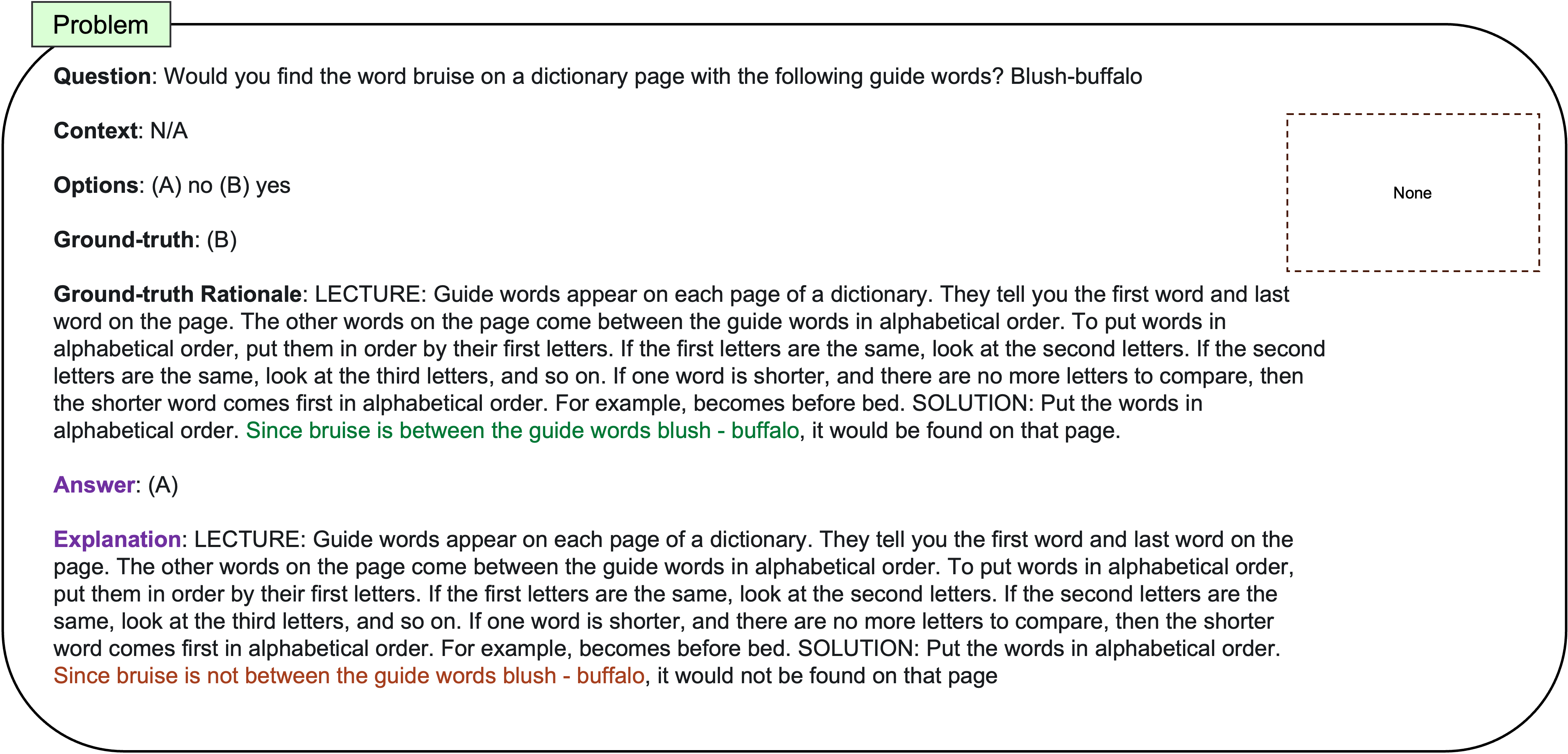}
   \caption{An example with incorrect answer and incorrect explanation}
   \label{fig:scitune_example_4}
\end{subfigure}
\caption{Two Unimodal QA examples with answer and explanation generated by LLaMA-SciTune}
\label{fig:scitune_example_unimodal}
\end{figure*}

Figures~\ref{fig:scitune_example_multimodal} and~\ref{fig:scitune_example_unimodal} show a few answers, lectures and solutions generated by LLaMA-SciTune (13B, CTOM) for ScienceQA test instances. 

